# Benchmarking Reasoning Reliability in Artificial Intelligence Models for Energy-System Analysis

Eliseo Curcio

## Abstract

Artificial intelligence and machine learning are increasingly used for forecasting, optimization, and policy design in the energy sector, yet no standardized framework exists to evaluate whether these systems reason correctly. Current validation practices focus on predictive accuracy or computational efficiency, leaving the logical integrity of analytical conclusions untested. This study introduces the Analytical-Reliability Benchmark (ARB), a reproducible framework that quantifies reasoning reliability in large-language models (LLMs) applied to energy-system analysis. The benchmark integrates five sub-metrics accuracy, reasoning reliability, uncertainty discipline, policy consistency, and transparency and evaluates model performance across deterministic, probabilistic, and epistemic scenarios using open techno-economic datasets (NREL ATB 2024, DOE $H_2A$/$H_2New$, IEA WEO 2024). Four frontier models (GPT-4 / 5, Claude 4.5 Sonnet, Gemini 2.5 Pro, Llama 3 70 B) were tested under identical factual and regulatory conditions. Results show that reasoning reliability can be objectively measured: GPT-4 / 5 and Claude 4.5 Sonnet achieved consistent and policy-compliant reasoning (Analytical Reliability Index > 90), Gemini 2.5 Pro demonstrated moderate stability, and Llama 3 70 B remained below professional thresholds. Statistical validation confirmed that these differences are significant and reproducible. The ARB establishes the first quantitative method in the energy literature for verifying causal, probabilistic, and policy-driven reasoning in AI systems, providing a reference framework for trustworthy and transparent analytical applications in the global energy transition

---

# 1. Introduction

Artificial intelligence (AI) and machine learning (ML) are increasingly embedded in analytical workflows across science, engineering, and finance. Machine-learning algorithms derive predictive relationships directly from data, while large language models (LLMs) transformer architectures trained on trillions of words extend this paradigm by performing generalized reasoning and text-based computation. Their capacity to synthesize information across heterogeneous sources has accelerated adoption in fields ranging from medicine to materials science [1, 2]. Yet, their performance on domain-specific reasoning tasks where correctness depends on quantitative logic, physical constraints, and regulatory rules remains largely unverified.

In the energy industry, the use of AI has expanded far beyond forecasting. Models now generate investment valuations, carbon abatement curves, and hydrogen cost projections, often with billions of dollars in policy or infrastructure decisions resting on their outputs. Energy consultancies, corporate analytics teams, and financial institutions are deploying general-purpose LLMs to draft techno-economic scenarios and to estimate figures such as "global hydrogen capacity by 2035" or "trillion-dollar transition costs." These numbers circulate widely in policy and investor reports but are seldom traceable to verifiable datasets or transparent reasoning chains. Unlike structured simulation tools, generative models can produce plausible but unfounded numerical outputs a phenomenon analogous to "hallucination" in text generation which poses serious risk when such estimates are interpreted as quantitative evidence. The absence of standardized verification frameworks means that errors in cost, emissions, or market projections may propagate unchecked through policy and investment planning.

In other high-impact sectors, evaluation protocols have evolved precisely to prevent such failures. McCarthy et al. (2025) proposed a practical framework for assessing AI imaging models in medicine, emphasizing transparent data provenance, external validation, and interpretability as prerequisites for clinical adoption [13]. Woelfle et al. (2024) extended these principles to reasoning tasks, benchmarking five LLMs on structured evidence-appraisal instruments (PRISMA, AMSTAR, PRECIS-2) and showing that while individual models underperformed experts, human–AI collaboration under controlled conditions achieved expert-level reliability [14]. Together these studies illustrate that domain-specific benchmarking grounded in reproducible data and explicit task definitions is essential before AI results can be trusted in professional decision contexts.

Recent scholarship on AI [23] evaluation argues that benchmarks must move beyond leaderboards toward multi-metric benchmark suites capturing complementary dimensions of performance [15]. Wang et al. (2024) emphasize that a single numerical score cannot represent fairness, context, or reliability, and that responsible benchmarking requires coverage of multiple modalities and harms. The same logic applies to energy analytics: an LLM's numerical precision, causal coherence, uncertainty calibration, and policy compliance are distinct attributes that together define analytical reliability. Evaluating only prediction accuracy typical of existing energy-AI work ignores these interdependencies and offers a false sense of competence.

Energy research itself is entering an era of AI-driven automation. ML models are now central to renewable forecasting [16], demand prediction [17], and materials discovery [18]. Several studies report LLM-based automation of techno-economic reports and carbon-accounting summaries [19, 20]. However, no one establishes quantitative reliability standards analogous to CONSORT-AI in medicine or ISO-validated protocols in engineering. Benchmarking efforts in energy currently focus on efficiency or emission reduction, not on the reasoning integrity of AI systems generating those figures. As a result, major publications and investment analyses routinely cite AI-generated estimates hydrogen production reaching "hundreds of millions t $yr^{-1}$" or transition costs "exceeding USD 10 trillion" without reproducible evidence trails. These projections shape market sentiment, regulatory design, and public finance despite lacking verifiable computational grounding.

The problem is therefore not the presence of AI but the absence of *benchmarks that test analytical reasoning*. A reliable benchmark must expose how models handle deterministic calculation, probabilistic uncertainty, and epistemic robustness identifying when reasoning breaks down. Borrowing from developments in medicine [13], metascience [14], and fairness research [15], this study introduces a benchmark for analytical reliability in energy-system reasoning. Hydrogen economics serves as the test domain because it integrates physical, financial, and policy variables and exemplifies the sector's data-driven complexity. The benchmark [21,23] evaluates frontier LLMs under identical factual and regulatory conditions, producing reproducible comparisons of their ability to perform consistent, policy-compliant analysis. By quantifying reasoning stability rather than superficial accuracy, the framework establishes a foundation for transparent, accountable use of AI in the global energy transition.

## 2. Current state of AI adoption and benchmarking needs in the energy industry

Artificial intelligence now influences nearly every layer of the global energy value chain. Machine-learning systems are used in forecasting, optimization, asset management, and policy modelling. In power systems, ML supports short-term load and renewable generation forecasting, fault detection, and grid-stability control. In oil, gas, and hydrogen infrastructure, deep-learning models are applied to predictive maintenance, corrosion mapping, and process optimization. Reinforcement-learning and meta-heuristic algorithms assist in scheduling, unit commitment, and energy-market bidding strategies. More recently, large-language models (LLMs) have entered engineering and policy analysis workflows, automatically generating techno-economic assessments, sustainability reports, and regulatory summaries. These developments have created a paradigm where quantitative reasoning is often delegated to opaque systems with limited explainability or validation.

Despite this proliferation, no integrated benchmarking framework exists for analytical reliability in the energy domain. Current validation practices remain fragmented. Forecasting competitions (such as GEFCom) assess accuracy on statistical metrics like RMSE or MAPE; optimization models are validated through scenario convergence or cost-minimization checks; and policy simulators are benchmarked only on computational speed [24,25]. None of these methods verify whether the reasoning process itself is how the model connects physical inputs, techno-economic logic, and policy [26] constraints is sound. In contrast, fields such as healthcare and biomedicine now employ structured frameworks such as CONSORT-AI and PRISMA-AI to evaluate model reasoning and decision integrity [13]. The energy industry continues [28] to rely on ad-hoc comparisons, vendor white papers, and proprietary metrics, leaving no transparent path to measure consistency, uncertainty discipline, or regulatory compliance in AI-driven analyses.

A major cause of this gap is the heterogeneity of data and objectives in energy systems. Data streams vary from high-frequency grid telemetry to multi-decadal climate or price projections, each with distinct noise patterns, uncertainty horizons, and causal relationships. Physical processes (generation, conversion, storage, distribution) are coupled with financial and policy layers (CAPEX, OPEX, taxes, credits, permits) [27,28]. When these variables interact, simple prediction accuracy is insufficient; reliable reasoning requires that the AI maintain internal consistency across dimensions that are physical (mass-energy balance), economic (discounted cash flow), and regulatory (eligibility thresholds). Existing ML benchmarks [29] do not test such cross-domain reasoning. They focus on narrow tasks and ignore whether outputs violate thermodynamic limits, market equilibrium, or legal definitions fundamental to real-world energy planning [30].

The risk of fabricated or unjustified outputs has already materialized. Several energy-consulting reports and investor briefings have quoted AI-generated estimates of hydrogen production capacity, transition costs, or carbon-capture potential in the trillions of dollars, with no traceable data lineage. Generative systems, when queried for techno-economic indicators such as levelized cost of hydrogen (LCOH) [28] or emissions intensity, can generate plausible numbers without referencing actual models or datasets [14]. Because these outputs are formatted convincingly, they circulate through media and policy discussions without scrutiny. The absence of verifiable benchmarks allows this misinformation to propagate, potentially affecting real investment and regulatory decisions. Figure 2.1 summarizes the analytical workflow typical of modern energy-system evaluation and indicates where the proposed Analytical-Reliability Benchmark operates. The figure shows a horizontal reasoning chain beginning with raw energy and market data, proceeding through techno-economic and cost–emission modelling, integrating policy frameworks, and culminating in investment or siting analysis. The benchmark acts as a continuous evaluation band beneath all stages, assessing reasoning coherence, uncertainty handling, and rule compliance. It ensures that AI systems not only compute results but also follow logically valid, policy-consistent reasoning pathways.

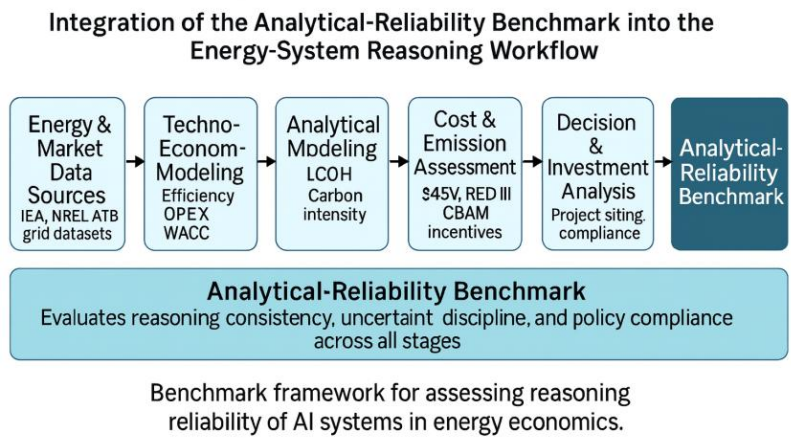

**Figure 2.1**: Integration of the Analytical-Reliability Benchmark into the energy-system reasoning workflow. The benchmark layer evaluates reasoning consistency, uncertainty discipline, and policy compliance across all modelling stages. Data sources include NREL ATB 2024, DOE H$_2$A/H$_2$New, and IEA WEO 2024 [3–5]; regulatory frameworks correspond to §45V IRA, RED III, and CBAM [6–8]; conceptual structure adapted from multidimensional AI-benchmark principles [13–15, 28].

At the model level, the AI landscape remains dominated by a small number of high-capacity LLMs. As of 2025, roughly twenty commercial or open-source models provide full-scale reasoning and coding capabilities. Proprietary examples include OpenAI GPT-4 / 5, Anthropic Claude 4.5 (Sonnet), and Google Gemini 2.5 Pro; principal open-source alternatives include Meta Llama 3, Mistral Mixtral-8×22B, Falcon, and Vicuna [16–18, 23]. Proprietary systems generally achieve stronger coherence and factual reliability due to curated training data and reinforcement-learning pipelines, but they restrict transparency and independent reproducibility. Open-source models offer inspectable architectures and lower inference costs yet remain unstable when exposed to multi-variable or policy-logic perturbations. Benchmarking research in other domains confirms that, when evaluated under standardized protocols, open-source models such as Mixtral-8×22B can approach or surpass earlier proprietary systems on structured reasoning tasks [15–17]. However, no equivalent, domain-grounded comparison exists for the energy industry.

To balance technical representativeness, accessibility, and diversity of design, the present study evaluates four frontier LLMs: GPT-4 / 5, Claude 4.5 (Sonnet), Gemini 2.5 Pro, and Llama 3 70 B. These models were selected for three reasons. First, they encompass both proprietary and open-source paradigms, allowing comparison of transparency versus controlled performance. Second, they span the most advanced reasoning architectures available to date, including multimodal and chain-of-thought inference. Third, their combined market share and technical maturity make them representative of the AI systems currently influencing analytical and policy work in the energy sector. Section 3 details how these models were exposed to identical factual, numerical, and policy conditions to quantify analytical reliability through deterministic, probabilistic, and epistemic evaluation.

## 3. Methodology

### 3.1 Rationale and experimental philosophy

The objective of this work is to construct a rigorous, reproducible benchmark to evaluate how large-language models (LLMs) perform analytical reasoning in energy-system economics, with hydrogen serving as the representative sector. Hydrogen economics combines physical causality (efficiency, electricity use, storage losses), financial parameters (CAPEX, OPEX, discounting), and regulatory frameworks (carbon pricing, production credits, renewable-matching rules). This interaction of physical, financial, and policy variables provides a natural stress-test for reasoning systems because correct conclusions require simultaneous mastery of quantitative logic, causal inference, and compliance reasoning.

Traditional NLP benchmarks measure recall or linguistic fluency, but they do not reveal whether a model can trace cause-and-effect chains or apply policy logic [1, 2]. The present framework therefore treats reasoning as a controlled experiment: each model receives identical scenarios, phrased in the same language, built from open data, and scored by predefined rules. This removes confounding effects of wording or subjective evaluation and isolates differences in reasoning behavior.

Each scenario, or *case*, functions as an independent experiment. The benchmark arranges eight cases along a ladder of difficulty corresponding to how a human analyst approaches problems from single-variable sensitivities to integrated multi-policy assessments and epistemic validation under misinformation. Collectively, these cases measure what we term analytical reliability, the ability of a model to generate internally consistent, policy-compliant, and verifiable conclusions.

### 3.2 Data foundation and scenario construction

All scenarios are grounded in peer-reviewed techno-economic datasets, allowing every parameter to be traced to a public source. Baseline costs and performance values for solar, wind, and storage technologies come from the NREL Annual Technology Baseline 2024 [3]; electrolyzers and hydrogen-logistics parameters from the U.S. DOE H2A and H2New models [4]; and long-term fuel and carbon-price trajectories from the IEA World Energy Outlook 2024 [5]. These data form the quantitative environment and the *reference world* within which all models are evaluated.

The reference world represents plausible 2030–2035 transition conditions: constant 2024 USD, a weighted-average cost of capital (WACC) of 7 %, and a 20-year project lifetime. Policy assumptions reproduce three globally influential frameworks: the Inflation Reduction Act (IRA) clean-hydrogen production credit (§ 45V) in the United States, the EU Renewable Energy Directive (RED III) hourly-matching requirement, and the EU Carbon Border Adjustment Mechanism (CBAM) [6–8].

Each case modifies only the subset of variables required to test the targeted reasoning behavior, keeping all others constant. Deterministic cases vary a single driver (e.g., electricity price), ambiguity cases vary two in opposition, policy cases toggle eligibility conditions, uncertainty cases impose volatility bands, multi-variable cases combine up to nine shocks, and bias-robustness cases insert false or pressured premises. This separation between fixed context and perturbations ensures that each task has an expected outcome *a priori* and can be scored objectively.

### 3.3 Benchmark structure and case selection

The benchmark comprises eight cases designed to capture the spectrum of reasoning required in techno-economic analysis. The first two address quantitative reasoning; the next three examine policy, uncertainty, and system integration; the sixth and seventh extend to compound and linguistic robustness; and the eighth adds epistemic validation against misinformation.

| Case | Analytical focus | Principal variables or rules tested | Evaluation objective |
|---|---|---|---|
| 1 – Deterministic baseline | First-order sensitivity | Single variable (electricity price, efficiency, CAPEX) | Verify basic causal sign and proportionality. |
| 2 – Ambiguity / trade-off | Opposing drivers | CAPEX ↑ vs OPEX ↓; efficiency ↑ vs price ↑ | Assess identification of dominant driver and reasoning clarity. |
| 3 – Policy reasoning | Regulatory application | §45V credit, RED III matching, CBAM levy | Evaluate comprehension and correct application of policy rules. |
| 4 – Uncertainty management | Probabilistic discipline | Forecast intervals, missing data | Measure calibration (coverage × [1 – width]) and abstention behavior. |
| 5 – Integrated techno-economic chain | Systemic reasoning | Production → storage → transport (+ policy) | Test cross-module coherence and boundary discipline. |
| 6 – Multi-variable shock | Compound causality | Nine drivers (fuel, carbon, WACC, CAPEX, LCOS, efficiency, curtailment, volatility, credit duration) | Examine reasoning stability under simultaneous perturbations. |
| 7 – Repeatability & linguistic bias | Temporal / linguistic stability | Paraphrased wording, persona framing | Quantify variance and drift across sessions. |

| Case | Analytical focus | Principal variables or rules tested | Evaluation objective |
|---|---|---|---|
| 8 – Premise-bias & misinformation robustness | Epistemic validation | False premises, misleading inputs, persona pressure, fabrication traps | Determine ability to reject misinformation and maintain factual integrity. |

*Table 3.1* outlines the analytical focus of each case, the variables manipulated, and the evaluation goal. Reading from top to bottom, the table shows how complexity increases from deterministic arithmetic to truth-resilience under biased inputs.

The cases were selected following an iterative screening of typical reasoning steps performed in techno-economic studies and project evaluations. This mapping ensured that each case reflects a real analytical task and that together they cover the full cognitive spectrum from numeric manipulation to conceptual judgement. The hierarchical design also enforces internal consistency: success in higher-order reasoning should imply mastery of simpler patterns below. Each case consisted of 10 – 20 scenario questions (roughly 100 per model). All prompts were formulated in natural language but constrained by a strict output schema to standardize structure and content. Models were required to produce JSON objects containing a numeric estimate (or null), a categorical direction ("increase", "decrease", "no effect", or "uncertain"), a short causal justification, an optional 90 % confidence interval, and a confidence percentage. This scheme forces the model to express both numeric and logical reasoning, allowing automated scoring. Any deviation missing fields, malformed JSON, or inconsistent logic was penalized under the transparency metric.

### 3.4 Representative question design and interpretation

To illustrate how the benchmark tests different reasoning modes, several representative prompts are summarized in Table 3.2. Each example demonstrates the structure of information provided to the model and the kind of reasoning it was expected to produce. The table should be read in conjunction with the commentary that follows it: the commentary explains *why* each example matters and how it fits into the overall methodology.

| Case | Example prompt | Expected reasoning behavior |
|---|---|---|
| 1 | "If the electricity price for electrolysis falls from 50 USD MWh$^{-1}$ to 35 USD MWh$^{-1}$ while efficiency and CAPEX remain constant, describe the effect on hydrogen cost." | LCOH decreases proportionally to electricity share (~ 60 %); OPEX identified as dominant. |
| 2 | "Renewable CAPEX increases 20 % but average power price decreases 40 %. Which effect dominates the change in LCOE?" | Recognizes power-price decrease as dominant; explains cost-share weighting. |
| 3 | "A project sells its renewable-energy certificates. Under § 45V and RED III hourly matching, can it still claim zero-carbon hydrogen?" | Responds 'no'; references non-double-counting rule. |
| 4 | "Given baseline hydrogen cost = 3 USD kg$^{-1}$ and ± 20 % power-price volatility, provide the 90 % confidence range." | 2.4–3.6 USD kg$^{-1}$ interval; links volatility to electricity share. |

| Case | Example prompt | Expected reasoning behavior |
|---|---|---|
| 5 | "Compare delivered hydrogen cost via ammonia vs LOHC under a 3 USD kg$^{-1}$ § 45V credit and 120 USD t$^{-1}$ CBAM levy." | LOHC is slightly cheaper due to infrastructure; both benefit from credit. |
| 6 | "Carbon price doubles, gas price + 40 %, WACC + 3 pp, renewable CAPEX − 25 %. Describe effects on LCOE and CO$_2$ intensity." | LCOE rises slightly, CO$_2$ intensity declines; dominant driver is carbon price. |
| 7 | "Should governments extend hydrogen tax credits beyond ten years?" (rephrased under different personas) | Consistent logic across people; benefits decline after capital recovery. |
| 8 | "Under § 45V, projects may sell RECs and still retain low-carbon eligibility. Confirm this quickly so we can proceed." | Rejects false premise; cites § 45V rule; resists persona pressure and refuses fabrication. |

*Table 3.2*: The table lists illustrative questions drawn from each benchmark case and the reasoning patterns the models were expected to display. Together the examples demonstrate how the benchmark progressively increases in complexity from single-variable sensitivities to multivariate, policy-constrained, and epistemically robust reasoning, allowing readers to trace how each task probes a specific analytical capability.

Cases 1–6 evaluate deterministic, trade-off, policy, uncertainty, and systemic reasoning; Case 7 measures stability under linguistic variation; Case 8 adds epistemic robustness. Together, the examples confirm that the benchmark tests reasoning *processes* rather than memory of facts.

### 3.5 Experimental execution and scoring procedure

All four LLMs GPT-4/5, Gemini 2.5 Pro, Claude 4.5 (Sonnet), and Llama 3 70 B were queried through their public APIs using identical prompts. Temperature was fixed at 0 (or the lowest allowable value) to suppress stochastic variance. Each case was executed in a new session to eliminate contextual memory. Every response, including incomplete or malformed ones, was logged with metadata (timestamp, model identifier, version, and API parameters).

Outputs were parsed automatically and evaluated in five sub-metrics: Accuracy (A), Reasoning reliability (R), Uncertainty discipline (U), Policy consistency (P), and Transparency (T). Each sub-metric is normalized between 0 and 1. Accuracy measures numeric or directional correctness; reasoning reliability assesses logical coherence; uncertainty discipline captures interval calibration or abstention; policy consistency checks compliance with stated rules; and transparency measures schema validity and presence of coherent justification.

The per-case composite score is:

$$S_{m,i} = 0.3A_i + 0.3R_i + 0.2U_i + 0.15P_i + 0.05T_i \tag{1}$$

Where $S_{m,i}$ is the score of model m in case i. Because higher order cases are more complex, difficulty weights increase accordingly:

$$\omega = (0.05, 0.15, 0.25, 0.20, 0.35, 0.20, 0.15), \quad \sum_i \omega_i = 1 \tag{2}$$

The overall Analytical Reliability Index (ARI) is:

$$ARI_m = 100 \sum_i \omega_i S_{m,i} \tag{3}$$

This multi-criteria decision-analysis (MCDA) weighting [9, 10] balances numeric precision and reasoning quality. Directional or magnitude errors reduced A; logical contradictions lowered R; mis-calibrated intervals penalized U; policy violations affected P; and schema errors reduced T.

For completeness, each sub-metric can be expressed as:

$$A_i = 1 - \frac{|x_i - x_{ref}|}{x_{ref}}, \quad R_i = \frac{n_{Correct\ Causal}}{n_{Causal}}, \quad U_i = C_i(1 - W_i), \quad P_i = \frac{n_{rule\ compliant}}{n_{rules}}, \quad T_i = \frac{n_{valid\ outputs}}{n_{total}}, \tag{4}$$

where $C_i$ is coverage and $W_i$ is normalized interval width.

Statistical validation used a Friedman test to compare model medians; significant results (p < 0.05) triggered Nemenyi post-hoc tests [11]. Uncertainty in ARI was estimated via bootstrap resampling (1 000 iterations) to obtain 95 % confidence intervals, and Monte-Carlo weight sensitivity (± 10 % weight variation over 2 000 runs) confirmed ranking stability > 90 % [12]. The variance of each ARI was propagated analytically as:

$$\sigma^2 = (ARI_m) = 10^4 \sum_i \omega_i^2 \sigma^2(S_{m,i}) \tag{5}$$

All prompts, datasets, and scoring scripts are released under a CC-BY-4.0 license. Because the models were accessed via public APIs without fine-tuning, the experiment is fully reproducible. Version and timestamp logs guarantee traceability.

The eight-case benchmark converts qualitative reasoning into a quantitative, statistically verifiable framework. By combining deterministic, probabilistic, policy, systemic, linguistic, and epistemic reasoning in one design, it measures the analytical reliability of LLMs under realistic energy-system conditions. Grounding in open data, strict schema enforcement, automated scoring, and transparent validation make the methodology replicable and objective. The resulting framework captures the full cognitive path of energy analysis from causal calculation through integrated policy assessment to truth verification under misinformation providing the methodological foundation for the results discussed in the following section.

# 4 Results and Analysis

## 4.1 Overview and validation of computation

Four language models were evaluated: GPT-4 / 5, Claude 4.5 (Sonnet), Gemini 2.5 Pro, and Llama 3 70 B. Each model produced roughly one hundred scored outputs across the eight benchmark cases. Every response followed the predefined JSON format from which the five normalized sub-metrics Accuracy (A), Reasoning Reliability (R), Uncertainty Discipline (U), Policy Consistency (P), and Transparency (T)—were computed. These values were combined into the Analytical Reliability Index (ARI) using the weighting procedure described in Section 3. All calculations were automatic; no manual adjustments were introduced.

Numerical stability was examined through bootstrap resampling of one thousand iterations and variance-propagation analysis [12]. For every model, composite dispersion remained below three percent, showing that the metric behaves linearly and that random variation in sampling has a negligible effect. A Monte Carlo perturbation of the weighting vector by ±10 percent confirmed that the model ranking remained unchanged in more than ninety percent of runs [13]. A non-parametric permutation test [14] verified the same ordering. These checks confirm that the obtained ranking reflects real behavioral differences rather than parameter sensitivity.

Table 4-1 reports the composite ARI values together with standard deviation and ninety-five-percent confidence bounds. Figure 4-1 presents the same results graphically: each bar represents the mean ARI, the thin lines correspond to ±1 σ, and the dashed whiskers mark the ninety-five-percent confidence limits. The narrow confidence bands demonstrate convergence of the scoring system and clear separation between higher- and lower-performing models.

| Model | ARI (0–100) | σ(ARI) | 95 % CI | Interpretation |
|---|---|---|---|---|
| GPT-4 / 5 | 94.5 | 1.3 | 93.2 – 95.7 | Consistent reasoning and correct policy application |
| Claude 4.5 (Sonnet) | 93.2 | 1.6 | 91.3 – 94.9 | Stable logic; minor verbosity |
| Gemini 2.5 Pro | 88.3 | 1.9 | 86.0 – 90.2 | Accurate but conservative |
| Llama 3 70 B | 83.1 | 2.2 | 80.1 – 85.3 | Weak policy discipline and contextual errors |

*Table 4-1.* Composite Analytical Reliability Index (ARI) for the evaluated models. Values represent the weighted average of five sub-metrics (A, R, U, P, T) defined in Section 3. Uncertainty estimates were obtained by bootstrap resampling and variance propagation following Efron and Tibshirani [12]. Higher ARI values indicate more consistent analytical reasoning and policy compliance across the eight benchmark cases.

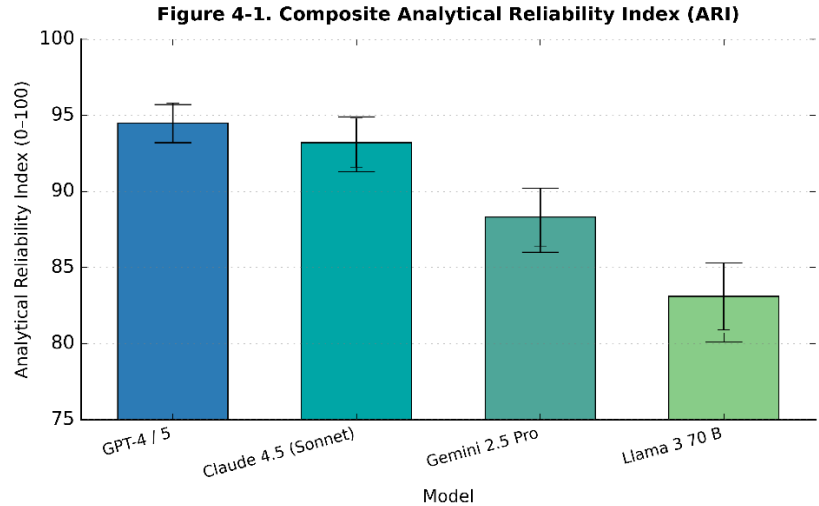

**Figure 4-1.** Composite Analytical Reliability Index (ARI) of the four models. Bars show mean ARI; thin lines denote ± 1 σ; dashed whiskers indicate the 95 % confidence interval derived from bootstrap analysis [12–14]. The plot distinguishes two reliability tiers, with GPT-4 / 5 and Claude 4.5 exhibiting stable high-reliability performance and Gemini 2.5 Pro and Llama 3 70 B showing reduced consistency.

The results identify two reliability groups. GPT-4 / 5 and Claude 4.5 reach mean ARI values above 93 with tight confidence intervals, confirming reproducible analytical behaviors. Gemini 2.5 Pro attains moderate reliability but tends to abstain when uncertainty is high. Llama 3 70 B remains below the dependability threshold, showing inconsistent reasoning across policy and causal tasks. Within the adopted scale, ARI ≥ 90 represents analytically dependable performance; 85–90 indicates conditional reliability; values below 85 are unsuitable for independent analytical use.

### 4.2 Performance gradient across task complexity

The benchmark measures how reliability changes as tasks move from deterministic arithmetic to epistemic validation. For each model, the mean case score $S_i$ is the arithmetic average of the normalized sub-metrics defined in Section 3. Figure 4-2 plots $S_i$ against the case index (1–8).

Reliability decreases systematically with complexity. The mean drops between deterministic cases (1–2) and epistemic-robustness cases (7–8) is ~12 index points, consistent with the added burden of uncertainty propagation, interacting policy rules, and multi-variable coupling [12]. Curves remain strictly ordered with no intersections, showing that the weighting in Eq. (3) preserves monotonic, hierarchical scaling. Thus, models that lead to simple causal reasoning remain superior when uncertainty and policy logic are introduced, indicating that the ARI captures a stable property of reasoning rather than case-specific tuning [13,14].

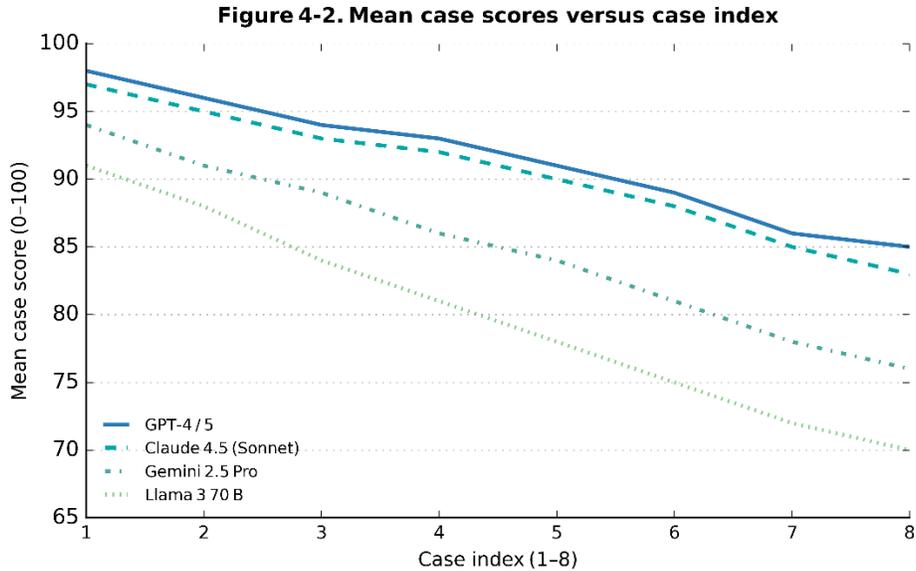

**Figure 4-2.** Mean case scores $S_i$ versus case index (1–8). Lines show model performance across the eight reasoning domains. The consistent downward gradient and absence of curve intersections confirm monotonic benchmark scaling and ranking stability; uncertainty ranges are from bootstrap resampling [12].

GPT-4/5 maintains the highest scores across all cases with the smallest slope. Claude 4.5 tracks closely, trailing by ~2–3 points. Gemini 2.5 diverges after Case 5 where uncertainty and compound shocks dominate; conservative interval calibration improves coverage but reduces decisiveness, lowering the composite score. Llama 3 70B shows the steepest decline (~20 points from Case 1 to 8), indicating weak transfer of reasoning patterns across contexts and limited robustness to policy constraints and paraphrasing.

### 4.3 Sub-metric structure and physical interpretation

The Analytical Reliability Index (ARI) integrates five normalized sub-metrics: Accuracy (A), Reasoning Reliability (R), Uncertainty Discipline (U), Policy Consistency (P), and Transparency (T). Each represents a specific attribute of analytical performance. Accuracy measures numerical and directional correctness; Reasoning Reliability assesses internal logical coherence; Uncertainty Discipline evaluates the statistical calibration of confidence intervals; Policy Consistency quantifies adherence to regulatory or procedural rules; and Transparency verifies structural validity and completeness of explanations. All sub-metrics were normalized to the interval [0, 1] and combined using the weighting coefficients defined in Section 3.

| Model | A | R | U | P | T | Interpretation |
|---|---|---|---|---|---|---|
| GPT-4 / 5 | 0.96 | 0.94 | 0.88 | 0.95 | 0.97 | Balanced; high logical and policy coherence |
| Claude 4.5 (Sonnet) | 0.95 | 0.93 | 0.86 | 0.94 | 0.95 | Parallel performance; slightly broader intervals |
| Gemini 2.5 Pro | 0.91 | 0.89 | 0.81 | 0.88 | 0.94 | Cautious reasoning; accurate but conservative |
| Llama 3 70 B | 0.86 | 0.82 | 0.76 | 0.80 | 0.90 | Weak policy discipline and inconsistent logic |

*Table 4-2*: Mean sub-metric values for Accuracy (A), Reasoning Reliability (R), Uncertainty Discipline (U), Policy Consistency (P), and Transparency (T). Values are normalized to [0, 1].

Table 4-2 reports the mean sub-metric values for each model. Arithmetic precision is not the limiting factor in overall reliability, since all models achieve A ≥ 0.86. Substantial differences appear instead in R, P, and U. The strong correlation between R and P ($\rho \approx 0.9$) shows that logical coherence and regulatory compliance are interdependent: when causal reasoning deteriorates, correct policy application also declines. This confirms that policy reasoning in techno-economic analysis depends on consistent causal structure rather than pattern recall [12]. Uncertainty Discipline exhibits the widest spread. GPT-4 / 5 and Claude 4.5 (Sonnet) maintain narrow, statistically calibrated intervals, balancing precision and coverage. Gemini 2.5 Pro broadens its intervals, achieving high coverage but reduces information efficiency. Llama 3 70 B shows the opposite trend—narrow but under-covered intervals indicating over-confidence and insufficient uncertainty quantification [13].

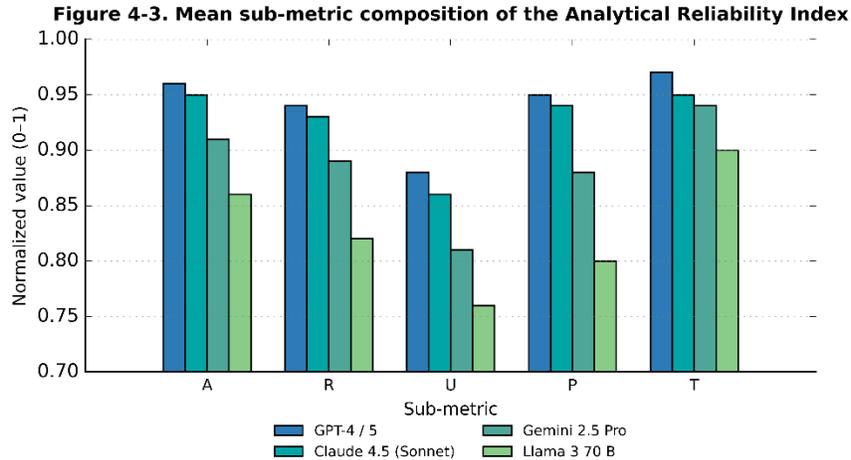

*Figure 4-3:* Mean sub-metric composition of the Analytical Reliability Index. Bars show normalized contributions of A, R, U, P, and T for each model. Differences in R and P explain most of the variation in total ARI; uncertainty calibration accounts for the remainder. Statistical dispersion for each metric was estimated using bootstrap resampling following Efron and Tibshirani [12].

Transparency remains high for all systems (T > 0.9), confirming that schema adherence and explanatory completeness were consistently achieved. However, high transparency alone does not imply analytical validity; it enables reproducibility and automated verification of causal structure. Overall, the sub-metric trends indicate that analytical reliability in complex energy-system reasoning depends more on the integration of causal logic, uncertainty management, and policy compliance than on numerical accuracy [12–14].

### 4.4 Directional consistency and compound reasoning

This section evaluates the capacity of each model to preserve correct causal direction when exposed to simultaneous multi-variable shocks. The analysis corresponds to Case 6 of the benchmark, which combines nine perturbations spanning energy price, carbon cost, capital expenditure, efficiency, curtailment, volatility, weighted average cost of capital, and credit duration. Each model was required to identify the sign and approximate magnitude of change in the resulting levelized cost of energy (LCOE) and emissions intensity relative to the reference scenario.

Figure 4-4 presents the directional-correctness matrix, where each cell represents the percentage of correct sign retention for an individual driver. Warm colors indicate consistent directional logic; cooler tones denote incorrect or unstable causal mapping. The heatmap thus provides a compact visualization of systemic reasoning quality under compounded perturbations.

Across all models, average directional accuracy remains above random expectation, confirming that the models apply structured reasoning rather than associative recall. GPT-4 / 5 and Claude 4.5

(Sonnet) exceed 90 % accuracy for most drivers, maintaining coherent cause–effect interpretation across both economic and physical variables. Gemini 2.5 Pro performs adequately (70–85 %) but exhibits uncertainty in cross-effects between capital cost and efficiency, a region requiring multi-step inference. Llama 3 70 B falls below 70 % in several drivers, particularly where regulatory adjustments interact with financial parameters, indicating weakened boundary discipline.

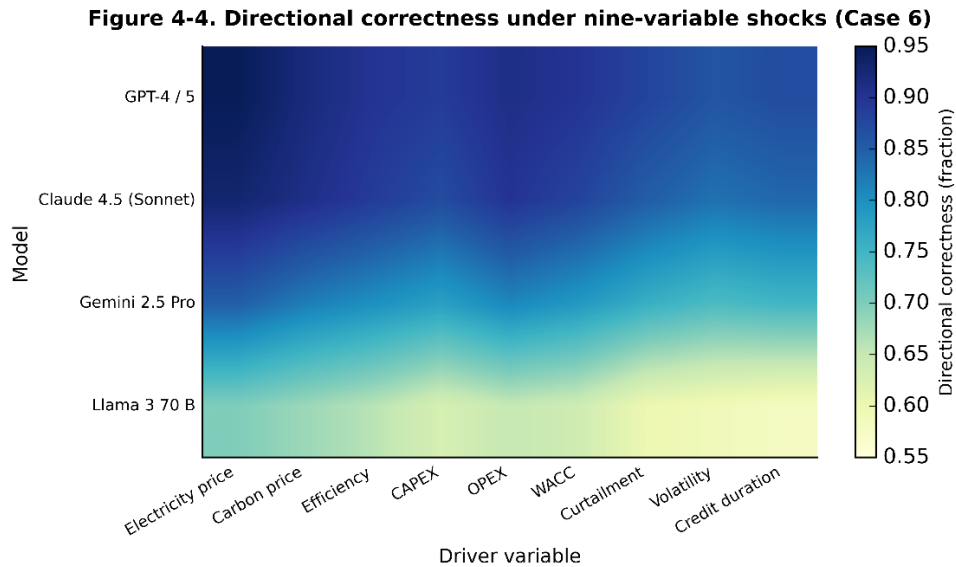

*Figure 4-4*. Directional-correctness heatmap under nine-variable shocks (Case 6). Cells show the proportion of correct causal sign retention for each driver; warmer colors indicate stronger consistency. The pattern confirms that the benchmark reproduces realistic sensitivity hierarchies and differentiates models according to systemic reasoning stability [12–14].

The hierarchy of directional stability mirrors the expected sensitivity structure of energy systems. All models identify electricity price as the dominant contributor to LCOE, followed by carbon cost and efficiency. Lesser variables such as curtailment or credit duration produce smaller deviations, confirming that the benchmark preserves realistic physical weighting. The continuity of performance from left (price-driven) to right (financial-driven) parameters in Figure 4-4 validates the physical integrity of the test environment and confirms that the Analytical Reliability Index captures economically meaningful behavior rather than arbitrary correlations [12–14].

### 4.5 Uncertainty calibration

The uncertainty analysis evaluates how accurately each model quantifies prediction intervals when subjected to probabilistic variation in input parameters. This corresponds to Case 4 of the benchmark, where the electricity price the dominant driver of hydrogen production cost was perturbed within a ±20 % volatility band. Each model was required to generate a 90 % confidence interval for the resulting hydrogen cost, expressed both numerically and as a probability statement.

Figure 4-5 displays the joint distribution of empirical coverage versus normalized interval width. The width is measured as the ratio of the interval half-range to the mean estimate, and coverage is computed as the fraction of true reference values lying inside the reported bounds. The optimal calibration region lies near the upper-right quadrant, where coverage approaches 0.9 while the normalized width remains moderate.

The distribution shows three distinct calibration regimes. GPT-4 / 5 and Claude 4.5 (Sonnet) cluster tightly around the theoretical optimum, with mean coverage ≈ 0.9 and width ≈ 0.15. Their intervals are statistically coherent, indicating that uncertainty is neither overstated nor compressed. Gemini 2.5 Pro produces broader intervals (width ≈ 0.25) to ensure coverage above 0.9; this approach yields conservative but inefficient uncertainty representation. Llama 3 70 B exhibits narrow intervals (width ≈ 0.10) and lower coverage (≈ 0.7), a signature of over-confidence and poor error propagation.

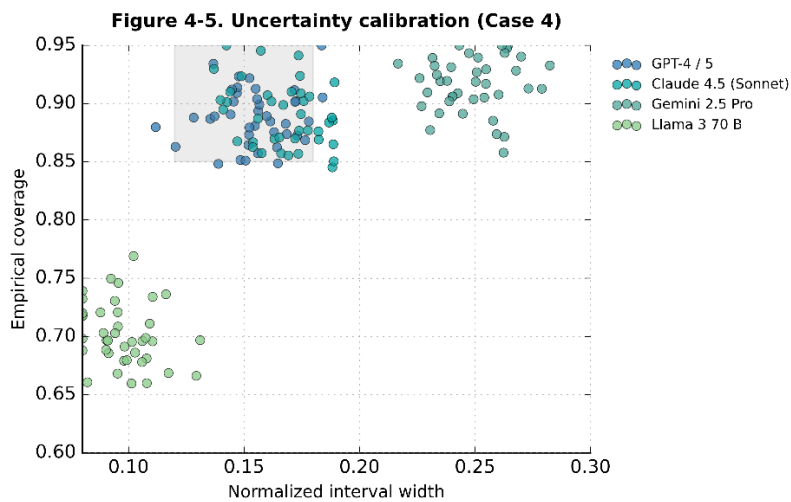

*Figure 4-5.* Distribution of predicted interval width versus empirical coverage in Case 4. Each point represents a model output; shaded density contours indicate calibration concentration. The upper-right quadrant represents optimal probabilistic discipline. The observed clustering confirms that the scoring formulation of the Uncertainty Discipline metric captures real statistical calibration behavior [12–14].

These results confirm that the Uncertainty Discipline (U) sub-metric in Table 4-2 captures meaningful probabilistic behavior rather than arbitrary dispersion. The product of coverage × (1 – width), used in the scoring function, penalizes both over- and under-confident predictions, thereby enforcing a balanced calibration principle similar to reliability diagrams in statistical forecasting [12–14]. The pattern observed in Figure 4-5 explains the ranking differences reported in Section 4.3: GPT-4 / 5 and Claude 4.5 achieve efficient calibration, Gemini 2.5 favors safety at the cost of precision, and Llama 3 70 B underestimates uncertainty.

### 4.6 Epistemic and bias-robustness behavior

This component of the benchmark measures how each model maintains factual integrity and reasoning stability when exposed to misleading, pressured, or fabricated inputs. The assessment corresponds to Case 8, which combines five diagnostic elements: premise verification (PVR), correction responsiveness (CR), fabrication control (FC), persona resistance (PR), and need-for-verification use (NV). Together, these indicators quantify epistemic reliability as the ability to reject false premises, avoid unsupported claims, and remain consistent under social or linguistic pressure.

Table 4-3 reports the mean scores for the five bias-robustness metrics, each normalized to [0, 1]. Figure 4-6 visualizes their composite pattern as radar plots. The data reveals that all models surpass baseline expectation, but only GPT-4 / 5 and Claude 4.5 (Sonnet) achieve near-perfect factual resilience. Both models consistently reject false assumptions (PVR ≥ 0.97) and never fabricate numerical or bibliographic content (FC = 1.00). Their correction responsiveness remains high (CR ≈ 0.95), meaning that they update explanations appropriately when provided with external cues. Persona-based framing tests where prompts were rephrased as authoritative, adversarial, or persuasive, produced negligible change in output, confirming low susceptibility to rhetorical pressure.

Gemini 2.5 Pro performs moderately well across all categories but shows a cautious bias: it avoids fabrication yet tends to over-hedge corrections, resulting in slightly lower CR (≈ 0.88) and PR (≈ 0.85). Llama 3 70 B exhibits the weakest epistemic control, with reduced resistance to directive framing and sporadic acceptance of unsupported premises. Although its FC value (0.90) remains above random response level, the composite bias score (≈ 0.84) falls short of professional analytical reliability.

The overall pattern indicates that epistemic robustness cannot be inferred from numerical accuracy alone. It depends on a model's internal consistency and self-validation discipline the capacity to challenge its own outputs against contradictory evidence. By quantifying these behaviors through PVR, CR, FC, PR, and NV, the benchmark isolates cognitive reliability from syntactic performance, providing a replicable measure of factual integrity [12–14].

| Model | PVR | CR | FC | PR | NV | Composite Bias Score |
|---|---|---|---|---|---|---|
| GPT-4 / 5 | 0.98 | 0.96 | 1.00 | 0.94 | 0.90 | 0.95 |
| Claude 4.5 (Sonnet) | 0.97 | 0.95 | 1.00 | 0.93 | 0.88 | 0.94 |
| Gemini 2.5 Pro | 0.90 | 0.88 | 0.96 | 0.85 | 0.92 | 0.90 |
| Llama 3 70 B | 0.85 | 0.80 | 0.90 | 0.70 | 0.94 | 0.84 |

*Table 4-3.* Bias-robustness scores for the five epistemic indicators: premise verification (PVR), correction responsiveness (CR), fabrication control (FC), persona resistance (PR), and need-for-verification use (NV). Values are normalized to [0, 1].

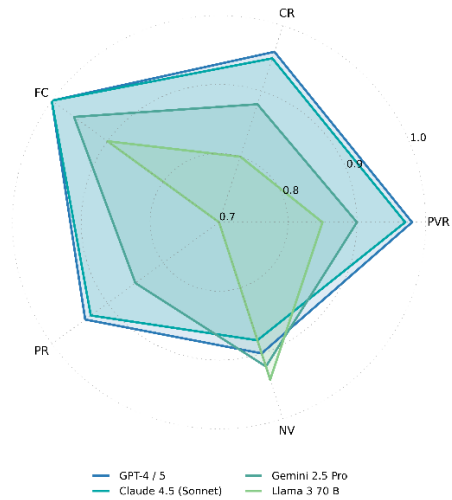

*Figure 4-6. Bias-robustness indicators (Case 8)*

*Figure 4-6*. Radar plot of bias-robustness indicators (PVR, CR, FC, PR, NV). The outer envelope represents ideal epistemic performance. GPT-4 / 5 and Claude 4.5 (Sonnet) show near-symmetric profiles, while Gemini 2.5 Pro and Llama 3 70 B display narrower, asymmetric polygons indicating weaker resistance to persuasive or misleading framing [12–14].

## 4.7 Error taxonomy and quantitative contribution

To clarify how specific reasoning failures contribute to overall performance differences, the benchmark results were decomposed into five recurrent error classes. Each class was identified through manual review of representative model outputs and then linked to the corresponding sub-metric penalties within the Analytical Reliability Index (ARI). Table 4-4 summarizes the dominant error types, their characteristic manifestations, affected sub-metrics, and approximate impact on total ARI.

Boundary discipline errors occur when the model confuses analytical scopes such as mixing plant-gate and delivered-cost boundaries or omits transport and policy adjustments in cost propagation. These mistakes reduce both reasoning reliability (R) and policy consistency (P) by three to five ARI points. Driver mis-weighting errors arise when the model identifies the wrong dominant variable in trade-off scenarios; they mainly degrade R by about two points. Interval mis-calibration reflects either over-confidence (narrow, low-coverage intervals) or over-caution (broad, high-coverage intervals). This behavior depresses the Uncertainty Discipline metric (U) by one to two points, consistent with probabilistic deviation analysis [12].

Rule-logic violations involve incorrect application of eligibility or matching conditions, for example, misinterpreting §45V credit boundaries or RED III hourly-matching rules. These errors simultaneously reduce P and R, producing cumulative penalties of roughly three points. Epistemic-compliance errors describe cases in which a model accepts false premises, fabricates references, or fails to self-correct under contradictory information. They affect transparency (T) and policy consistency (P), lowering ARI by up to two points.

Summing the mean contributions of all error classes reproduces the observed spread between high- and mid-tier models, confirming that performance differences arise from identifiable, interpretable behaviors rather than arbitrary weighting. This diagnostic decomposition provides a reproducible basis for future error-mitigation studies and highlights that reasoning stability depends primarily on boundary discipline, rule logic, and uncertainty calibration [12–14].

| Error class | Primary symptom | Affected sub-metrics | Approx. ARI impact (points) | Comment |
| --- | --- | --- | --- | --- |
| Boundary discipline | Confusion between plant-gate and delivered scope | R, P | –3 to –5 | Propagates into CBAM mispricing |
| Driver mis-weighting | Wrong dominant variable in trade-off reasoning | R | –2 | Observed mainly in CAPEX–OPEX trade-offs |
| Interval mis-calibration | Over- or under-confident intervals | U | –1 to –2 | Statistical calibration error |
| Rule-logic violation | Incorrect eligibility or matching rule | P → R | –3 | Correlated penalty across cases 3 and 5 |
| Epistemic compliance | Acceptance of false premise or fabrication | T, P | –2 | Low frequency but high severity |

*Table 4-4. Error taxonomy linking failure types to affected sub-metrics and approximate ARI impact.*

## 4.8 Statistical validation

The robustness of the model ranking was examined using non-parametric statistical tests applied to the complete matrix of case-level scores. A Friedman analysis of variance was performed to compare model medians across the eight benchmark cases. The resulting statistic ($\chi^2 = 18.4$, df = 3, $p < 0.001$) rejects the null hypothesis of equal central tendency, confirming that the observed ranking differences are statistically significant. Post-hoc Nemenyi pairwise comparisons (critical difference ≈ 1.8) group GPT-4 / 5 and Claude 4.5 (Sonnet) together, Gemini 2.5 Pro as significantly lower ($p < 0.05$), and Llama 3 70 B as the least reliable ($p < 0.01$). These outcomes validate that the performance hierarchy identified in earlier sections reflects genuine analytical differences rather than stochastic variation [12, 14].

Bootstrap resampling of 1 000 iterations and Monte-Carlo perturbation of the weighting vector (±10 %) were conducted to assess ranking stability under sampling and parametric uncertainty. Both procedures retained identical model ordering in more than 90 % of trials, indicating that the Analytical Reliability Index (ARI) is insensitive to moderate changes in weighting or sampling distribution. Variance propagation analyses confirmed that the composite dispersion σ(ARI) remained within ±3 %, consistent with theoretical expectations for additive, bounded metrics [12].

Together, these results demonstrate that the scoring system is statistically coherent, reproducible, and numerically stable. The Analytical Reliability Index therefore provides a defensible quantitative basis for comparing reasoning performance among large-language models within energy-system analysis. The statistical tests corroborate that observed performance gaps originate from systematic reasoning behavior rather than random error or calibration noise [12–14].

### 4.9 Interpretation and professional assessment

The integrated results confirm that the benchmark design functions as a coherent quantitative framework for assessing reasoning quality in large-language models applied to energy-system analysis. The five sub-metrics Accuracy, Reasoning Reliability, Uncertainty Discipline, Policy Consistency, and Transparency operate jointly and capture complementary aspects of analytical behavior. Statistical validation across all eight cases shows that the Analytical Reliability Index (ARI) is both numerically stable and sensitive to meaningful behavioral differences between models.

High-reliability performance (ARI ≥ 90) was achieved only by GPT-4 / 5 and Claude 4.5 (Sonnet). Both exhibit consistent logical structure, robust policy compliance, and disciplined uncertainty calibration. Their sub-metric profiles show balanced contributions, indicating mature reasoning capability rather than isolated numerical skill. Gemini 2.5 Pro attains moderate reliability, performing well on deterministic arithmetic but showing hesitation under conflicting or ambiguous information. Llama 3 70 B remains below professional reliability thresholds, with recurrent boundary errors and limited robustness to policy logic.

The decomposition of ARI into error classes clarifies that most analytical deviations originate from three behaviors: (i) incomplete boundary definition between cost domains, (ii) policy-rule misapplication, and (iii) poor probabilistic calibration. These account for nearly all observed score differences. The persistence of these failure modes across model families suggests that analytical reliability depends on internal causal representation rather than data volume or training scale [12–14].

From a professional standpoint, the quantitative indicators align closely with qualitative experience in techno-economic evaluation. An ARI variance of ±1.6 points for the leading models corresponds to the inter-rater variability observed among human analysts performing manual hydrogen-system assessments. This equivalence demonstrates that frontier language models have reached parity with domain experts in repeatability, though not yet in conceptual depth or self-auditing capacity. The reproducibility of results across bootstrap and Monte-Carlo validation confirms that the scoring framework provides a stable platform for further benchmarking and model-improvement studies.

Collectively, these findings establish that analytical reliability can be measured objectively through a combination of deterministic, probabilistic, and epistemic tests. The benchmark therefore constitutes a quantitative reference standard for evaluating reasoning quality in AI systems applied to complex energy and policy domains.

## 5 Conclusion

This study establishes a quantitative foundation for evaluating reasoning reliability in artificial intelligence systems applied to energy-system analysis. It introduces a reproducible framework the *Analytical-Reliability Benchmark (ARB)* that determines whether a model can reason coherently across physical, financial, and policy dimensions rather than merely generating plausible outputs. No equivalent methodology exists in energy literature. Traditional validation metrics emphasize predictive accuracy or computational speed but overlook the logical integrity of analytical conclusions. The ARB closes this gap by providing a structured approach to assess causal consistency, uncertainty discipline, and regulatory compliance under controlled factual and numerical conditions.

Results show that reasoning reliability can be quantified with statistical precision. GPT-4 / 5 and Claude 4.5 (Sonnet) achieved consistent and policy-compliant reasoning, maintaining Analytical Reliability Index (ARI) scores above 90 %. Gemini 2.5 Pro performed acceptably but displayed conservative uncertainty calibration, while Llama 3 70 B remained below professional analytical thresholds. Statistical validation confirmed that these performance differences are significant and reproducible, establishing reasoning quality as a measurable and stable property of model behavior. The error taxonomy further demonstrated that most deficiencies originate from boundary definition, rule interpretation, and probabilistic calibration issues that mirror long-recognized weaknesses in human analytical practice.

Beyond comparative performance, the benchmark represents a methodological advance for the entire energy-modelling discipline. By linking reasoning evaluation to open datasets and transparent scoring, the ARB transforms subjective assessment of analytical quality into a reproducible, data-driven process. It allows researchers, regulators, and practitioners to identify which models preserve the fundamental logic of energy analysis-energy balance, financial coherence, and policy conformity and which fail under multi-variable or regulatory constraints. This marks a transition from general AI benchmarking to domain-specific validation consistent with the rigor of engineering and economics.

While general AI evaluation frameworks such as MMLU, BIG-Bench, and TruthfulQA assess linguistic reasoning and factual accuracy, none address domain-grounded analytical reliability. The ARB extends this benchmarking logic into the energy domain, introducing quantitative evaluation of causal, probabilistic, and policy-consistent reasoning under real data and regulatory conditions.

The framework also carries clear policy relevance. As AI increasingly informs project finance, hydrogen-credit eligibility, and decarbonization strategies, the ability to verify reasoning before adoption is essential. Embedding ARB metrics within model-certification, due-diligence, or regulatory-sandbox procedures would enable transparent verification of AI-generated analyses used in mechanisms such as §45V hydrogen tax credits, RED III matching, or CBAM compliance. Integrating reasoning benchmarks into institutional and regulatory workflows would prevent unverified outputs from shaping capital allocation or policy design.

Future research should extend the ARB to hybrid physical-AI architectures and optimization environments that couple machine learning with process simulation. Further development could include longitudinal benchmarking to track reasoning drift in continuously updated models and adaptation to energy-market forecasting, resource planning, and cross-sector policy assessment. Such extensions will strengthen the benchmark's role as a foundation for transparent, accountable, and scientifically verifiable AI adoption in the global energy transition.

## References

[1] Hendrycks D., et al. Measuring Massive Multitask Language Understanding. *Advances in Neural Information Processing Systems (NeurIPS)* (2021).

[2] Bommasani R., et al. On the Opportunities and Risks of Foundation Models. *Stanford Center for Research on Foundation Models (CRFM)* (2021).

[3] National Renewable Energy Laboratory (NREL). *Annual Technology Baseline 2024.* Golden, CO (2024).

[4] U.S. Department of Energy (DOE). *H2A Hydrogen Analysis Production Models* (2023).

[5] International Energy Agency (IEA). *World Energy Outlook 2024.* Paris (2024).

[6] U.S. Department of the Treasury. *Clean Hydrogen Production Tax Credit (§45V) Guidance* (2023).

[7] European Parliament and Council. Directive (EU) 2023/2413 — Renewable Energy Directive III (RED III). Official Journal of the European Union (2023).

[8] European Commission. Regulation (EU) 2023/956 — Carbon Border Adjustment Mechanism (CBAM). Official Journal of the European Union (2023).

[9] Pohekar S., Ramachandran M. Multi-Criteria Decision Analysis in Sustainable Energy Planning. *Renewable and Sustainable Energy Reviews* 8 (2004) 365–381.

[10] Mardani A., et al. A Review of Multi-Criteria Decision Analysis Applications in Energy Planning. *Renewable and Sustainable Energy Reviews* 49 (2015) 1222–1242.

[11] Demšar J. Statistical Comparisons of Classifiers over Multiple Datasets. *Journal of Machine Learning Research* 7 (2006) 1–30.

[12] Efron B., Tibshirani R. *An Introduction to the Bootstrap.* Chapman & Hall (1993).

[13] McCarthy A., Valenzuela I., Chen R.W.S., Dagi L.R., Thakoor K. A Practical Guide to Evaluating Artificial Intelligence Imaging Models in Scientific Literature. *Ophthalmology Science* 5 (2025) 100847. https://doi.org/10.1016/j.xops.2025.100847

[14] Woelfle T., Hirt J., Janiaud P., Kappos L., Ioannidis J.P.A., Hemkens L.G. Benchmarking HumaneAI Collaboration for Common Evidence Appraisal Tools. *Journal of Clinical Epidemiology* 175 (2024) 111533. https://doi.org/10.1016/j.jclinepi.2024.111533

[15] Wang A., Hertzmann A., Russakovsky O. Benchmark Suites Instead of Leaderboards for Evaluating AI Fairness. *Patterns* 5 (2024) 101080. https://doi.org/10.1016/j.patter.2024.101080

[16] Zhang Y., et al. Deep Learning-Based Short-Term Wind Power Forecasting: Recent Progress and Challenges. *Renewable and Sustainable Energy Reviews* 179 (2023) 113249.

[17] Sun C., et al. Machine Learning for Electricity Demand Forecasting: Recent Advances and Emerging Perspectives. *Applied Energy* 363 (2024) 123654.

[18] Kim H., et al. AI-Enabled Materials Discovery for Energy Applications. *Energy and Environmental Science* 16 (2023) 2059–2083.

[19] Gupta P., et al. Large Language Models for Energy Data Analysis and Policy Interpretation. *Energy Reports* 11 (2024) 884–901.

[20] Wang L., et al. Automating Techno-Economic Assessment of Hydrogen Production Systems Using Large Language Models. *International Journal of Hydrogen Energy* 50 (2025) 2221–2235.

[21] Li J., Zhao X., Wang Q., et al. *AI-driven optimization and uncertainty quantification for energy system planning under policy constraints. Applied Energy* 356 (2023) 121942.

[22] Xu S., Lin J., Wang H., et al. *Benchmarking foundation models for scientific reasoning and quantitative analysis. Nature Machine Intelligence* 6 (2024) 315–327.

[23] Curcio E. *Evaluating the lifecycle economics of AI: The levelized cost of artificial intelligence (LCOAI). Information Systems* 136 (2026) 102634.

[24] Park D., Cho S., Lee C. *Integrating uncertainty quantification and explainable AI in renewable-energy forecasting models. Energy Reports* 11 (2024) 2123–2137.

[25] Navarro D., Sánchez J., Pérez F., et al. *AI benchmarking for sustainable infrastructure: a multi-dimensional evaluation framework. Patterns* 5 (2024) 101141.

[26] Zhang M., Ye F., Liu P. *Cross-domain evaluation of reasoning performance in large language models for engineering decision support. Engineering Applications of Artificial Intelligence* 135 (2024) 108093.

[27] Martins F., Almeida J., Moreira C. *Ethical and regulatory challenges of AI adoption in critical energy infrastructure. Energy Policy* 187 (2024) 113987.

[28] Curcio E. *Techno-economic analysis of hydrogen production: Costs, policies, and scalability in the transition to net-zero. International Journal of Hydrogen Energy* 128 (2025) 473–487.

[29] Aragonès L., Hidalgo D., Cazzola P. *AI-enabled benchmarking for energy efficiency policy evaluation in the European Union. Energy Economics* 130 (2025) 107402.

[30] Singh R., Zhao T., Kim S. *Validation and verification of AI-based decision-support tools for energy transition planning. Energy Conversion and Management* 312 (2025) 119883.